\documentclass[runningheads]{llncs}
\usepackage{amsmath}
\usepackage{amssymb}
\usepackage{booktabs}
\usepackage{diagbox}
\usepackage{multirow}
\usepackage[T1]{fontenc}
\usepackage{graphicx,verbatim}
\usepackage{booktabs}
\usepackage{multirow}
\usepackage{graphicx}
\usepackage[table,xcdraw]{xcolor}
\usepackage{color}
\usepackage{arydshln}
\newcommand{\datasetname}[1]{\textsc{#1}}
\newcommand{\modelname}[1]{\textsc{#1}}
\newcommand*{\dataset}{\datasetname{Scale-VQA}}
\newcommand*{\model}{\modelname{ScaleReasoner-R1}}

\newcommand*{\mcq}{4{,}685} 
\newcommand*{\rois}{2{,}537} 
\newcommand*{\wsis}{177} 
\newcommand*{\triplets}{937} 
\definecolor{pinky}{HTML}{FFE2E1}
\definecolor{aliceblue}{rgb}{0.94, 0.97, 1.0}

\begin{document}
\title{Enhancing Pathological VLMs with \\ Cross-scale Reasoning}
\titlerunning{Enhancing Pathological VLMs with Cross-scale Reasoning}

\author{
Chi Phan \inst{1}$^\star$
\and
Tianyi Zhang \inst{1}$^\star$
\and
Qiaochu Xue \inst{1} 
\and
Yufeng Wu\inst{2} 
\and
Dan Hu\inst{3} 
\and
Zeyu Liu\inst{2}
\and
Sudong Wang\inst{2}$^{\star\star}$
\and
Yueming Jin\inst{1}$^{\star\star}$
}

\authorrunning{C. Phan et al.}

\institute{
Department of Electrical and Computer Engineering, National University of Singapore, Singapore\\
\email{\{chiphan,zhangtianyi,e1352520\}@u.nus.edu, ymjin@nus.edu.sg}
\and
PuzzleLogic Pte Ltd, Singapore\\
\email{\{yufengwu,zeyuliu,sudongwang\}@puzzlelogic.com}
\and
Department of Pathology, Fujian Medical University Cancer Hospital \& Fujian Cancer Hospital, Fuzhou, China\\
\email{hudan@fjmu.edu.cn}
}
  
\maketitle
\begingroup
\renewcommand{\thefootnote}{\fnsymbol{footnote}}
\footnotetext[1]{Chi Phan and Tianyi Zhang contributed equally to this work.}
\footnotetext[2]{Sudong Wang and Yueming Jin are co-corresponding authors. }

\endgroup
\begin{abstract}
Pathological images are inherently multi-scale, requiring pathologists to integrate evidence from global tissue architecture at low magnification to cellular morphology at higher magnification for accurate diagnosis. While existing pathological datasets for vision-language models (VLMs) include various scales, they often lack explicit cross-scale reasoning objectives. This limitation prevents VLMs from capturing essential cross-scale representations and learning evidence-based reasoning. 
To bridge this gap, we introduce the first cross-scale training and evaluation paradigm that formulates pathology interpretation as multi-magnification reasoning. However, creating such a task reveals a critical challenge: multi-image visual question answering (VQA) is prone to text-only shortcuts, which allow models to guess answers using magnification-dependent artifacts rather than visual evidence. To address this, we propose a leakage-aware curation pipeline that combines adversarial text-only screening with constraint-guided question design. Using this pipeline, we construct \dataset, a high-quality benchmark with \mcq\ multiple-choice questions grounded in \rois\ pathology images across multiple magnification levels. 
Finally, we present \model, a model trained via reinforcement learning to optimize performance on cross-scale VQA tasks. 
\model\ achieves state-of-the-art performance on our cross-scale reasoning benchmark and generalizes to SOTA performance on established single-scale benchmarks.
Findings suggest that even the limited cross-scale supervision can significantly improve pathological understanding. Code is available at \url{https://github.com/iMVR-PL/ScaleReasoner-R1}.

\end{abstract}    
\section{Introduction}

Pathological image analysis is a core component of cancer diagnosis. With the increasing availability of digitized whole-slide images (WSIs), deep learning methods for computational pathology have advanced rapidly. Recent vision-language models (VLMs) have achieved strong performance on visual question answering (VQA) tasks in medical imaging~\cite{llavamed,lingshu,huatuogpt} and pathology~\cite{clover,quilt,pathor1}, enabling more interpretable and interactive computer-assisted workflows.

In routine practice, pathologists reason across magnifications~\cite{patho1,patho3,patho2}. At low magnification, they assess global tissue architecture (e.g., histologic subtypes and growth patterns). At high magnification, they examine cellular morphology (e.g., nuclear atypia and mitotic activity). Observations at one magnification often determine what to verify at another, and the final diagnosis emerges from integrating evidence across scales.

\begin{figure}[t]
    \centering
    \includegraphics[width=1\linewidth]{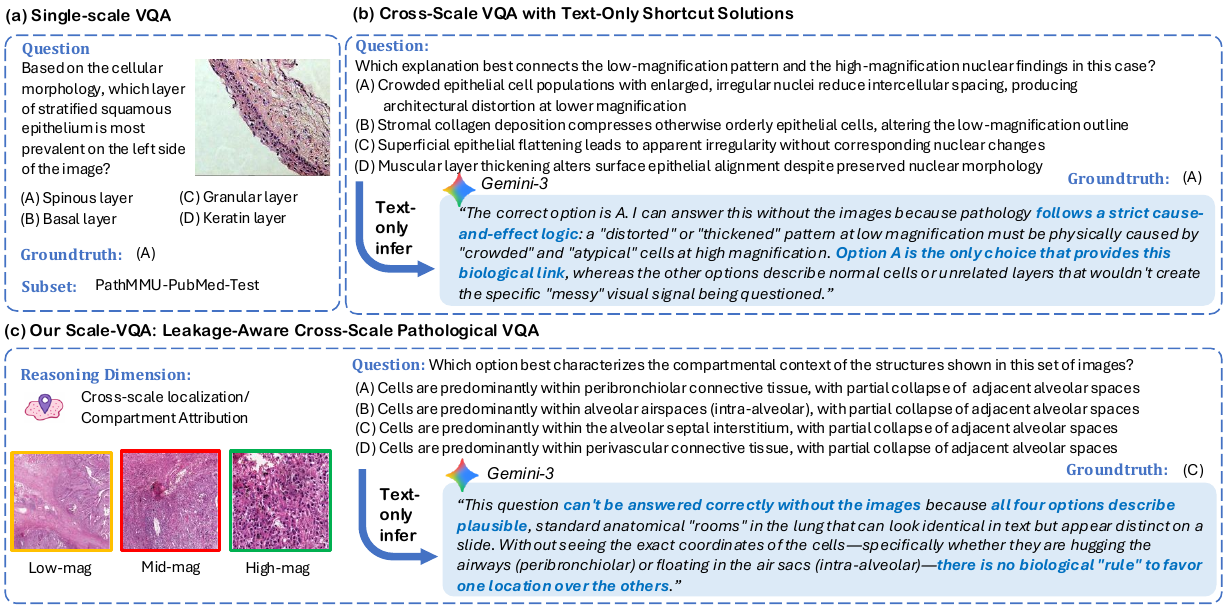}
    \caption{A comparison of (a) single-scale VQA, (b) naïve cross-scale VQA with text-only shortcut solutions, and (c) our leakage-aware cross-scale VQA.}
    \label{fig:examples}
\end{figure}

However, most existing pathology VLMs ~\cite{clover,quilt,smartpath,pathor1,slidechat,pathchat} are developed and evaluated on single-image tasks at a fixed magnification (Fig.~\ref{fig:examples}a). 
Correspondingly, many benchmarks treat isolated regions of interest (ROIs) as independent diagnostic units~\cite{pathvqa,pathmmu}, or represent WSI as thumbnails or feature embeddings at one specific scale~\cite{wsivqa,slidebench,wsibench}. 
This single-scale framing creates a mismatch with clinical practice: models may learn scale-specific recognition without acquiring the hierarchical rationale that links low-magnification architecture to high-magnification cellular evidence. 
Consequently, they often struggle with tasks that require cross-scale verification, such as localizing anatomical compartments (e.g., determining whether malignant cells invade specific tissue layers~\cite{slidechat}) or confirming suspected architectural patterns (e.g., validating a low-power suspicion of malignancy with high-power cellular findings~\cite{pathchat}). 
Bridging this gap requires moving beyond isolated recognition toward coordinated evidence integration across magnifications.

Formulating cross-scale VQA tasks, however, encounters a key challenge: text-only shortcut solutions~\cite{Agrawal_2018_CVPR,Goyal_2017_CVPR}. 
We refer to a shortcut solution as a case where the correct option can be inferred from the question and options without using the image evidence from all required magnification views~\cite{Agrawal_2018_CVPR,Goyal_2017_CVPR,shrestha-etal-2020-negative}. 
In pathology, this leakage can arise because some findings are strongly associated with specific magnifications, and option sets can contain systematic linguistic or biomedical priors (Fig.~\ref{fig:examples}b). 
Consequently, a naively constructed multi-image VQA benchmark may allow models to obtain high accuracy while ignoring one or more magnification views, leading to inflated results~\cite{Agrawal_2018_CVPR} that do not reflect genuine cross-scale evidence integration.
Therefore, reliable cross-scale training and evaluation requires leakage-aware benchmark construction, where questions are designed so that answering correctly depends on evidence across magnifications and text-driven shortcuts are explicitly suppressed~\cite{shrestha-etal-2020-negative}.

\begin{figure}[t!]
    \centering
    \includegraphics[width=1\linewidth]{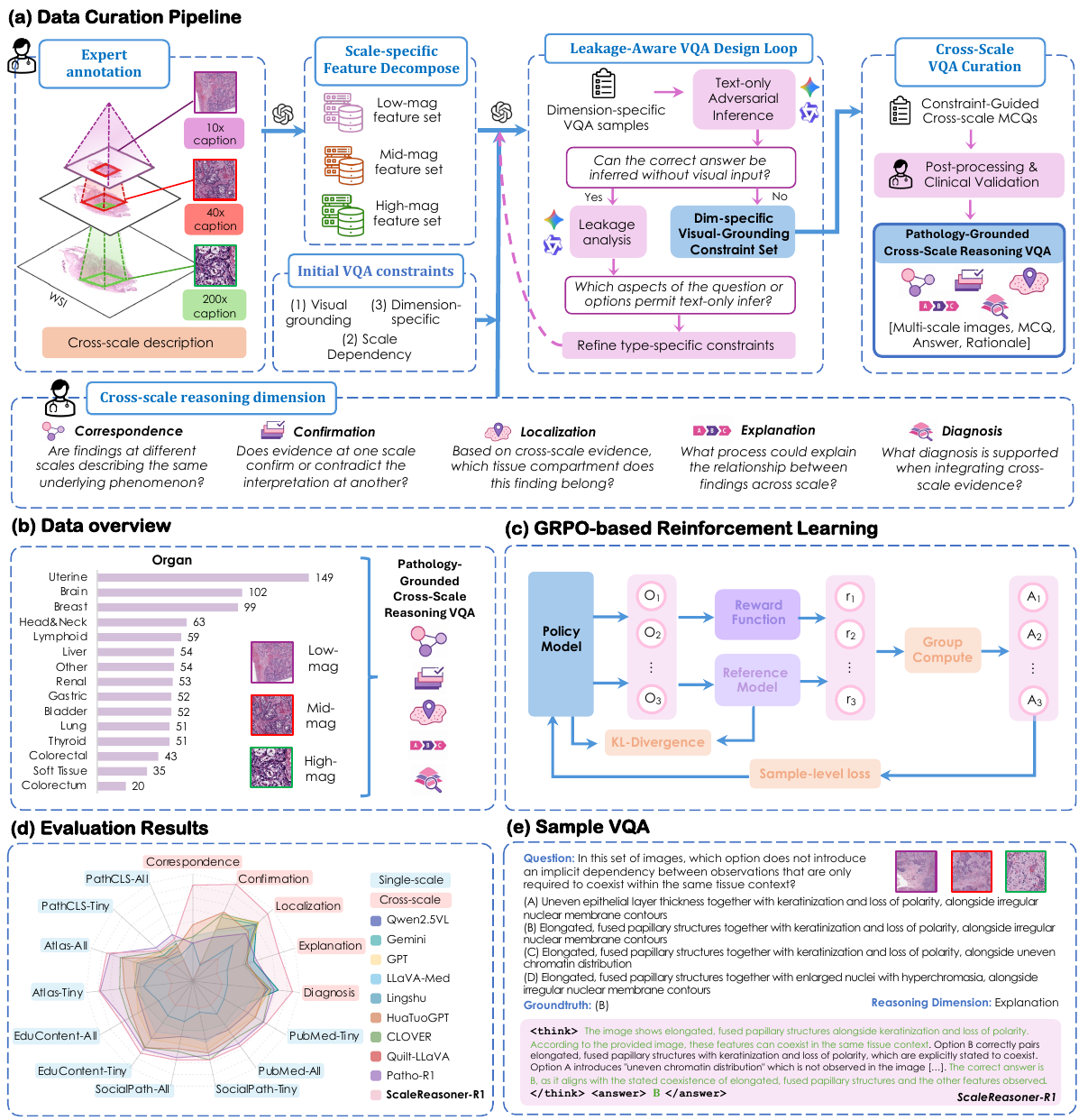}
    \caption{\textbf{Overview of \dataset{} and \model.} (a) Leakage-aware curation pipeline. (b) Dataset overview. (c) GRPO-based RL training. (d) Cross-scale reasoning results. (e) Example cross-scale VQA.}
    \label{fig:main}
\end{figure}

In this work, we introduce a cross-scale training and evaluation paradigm that frames pathology interpretation as coordinated reasoning across magnifications, rather than isolated single-scale recognition.
First, we develop a rigorous annotation pipeline: fifteen pathologists (>5 years of experience) and five senior pathologists (>20 years of experience) annotate \wsis\ publicly available TCGA WSIs~\cite{tcga}. We curate \rois\ ROI captions at 10\(\times\), 40\(\times\), and 200\(\times\), and collect \triplets\ cross-scale captions that synthesize visual patterns across magnifications.
Second, we design a leakage-aware VQA curation pipeline (Fig.~\ref{fig:main}a) that combines adversarial text-only screening with constraint-guided question design to suppress shortcut solutions and ensure each question depends on visual evidence across views. This produces \dataset, a dataset of \mcq\ multiple-choice questions that require visually grounded cross-scale reasoning.
Finally, we present \model, a model trained via curated reinforcement learning tasks. We enable the cross-scale reasoning insights to improve single-scale performance: \model\ achieves state-of-the-art (SOTA) performance on cross-scale multi-image benchmarks as well as existing single-image pathology VQA benchmarks.

In summary, we (i) introduce the first cross-scale paradigm for pathological reasoning, (ii) develop a high-quality cross-scale benchmark \dataset\ with a leakage-aware VQA curation pipeline, and (iii) release \model, a high-performing RL-trained model optimized for cross-scale VQA. Built on small-scale (\wsis\ WSIs) but high-quality data, we demonstrate that curated RL tasks can achieve state-of-the-art results on cross-scale VQA while transferring strongly to single-scale pathology VQA benchmarks.
Findings suggest that cross-scale supervision promotes more robust and interpretable pathological understanding even for single-scale perception. Underscoring its practical value for pathological image analysis, we open-source the high-quality \dataset\ and the curation pipeline to encourage broad community participation.

\section{Methods}
\label{sec:methods}
\subsection{Clinical Annotation for Cross-scale Reasoning}
We construct a clinically verified foundation for cross-scale reasoning by annotating \wsis\ publicly available TCGA WSIs~\cite{tcga}. 
Unlike prior benchmarks~\cite{pathmmu,wsibench,slidebench} that typically only involve experts at the final evaluation stage, our design integrates clinical expertise at the earliest stages of data curation to provide high-fidelity visual grounding and professional captioning.
To maximize diagnostic rigor, we utilize a collaboration workflow: each case was independently annotated by three junior pathologists and subsequently reviewed and validated by a senior pathologist. For every diagnostic case, annotators first select representative ROIs at low magnification (10$\times$) and further zoom in to pick corresponding regions at intermediate (40$\times$) and high (200$\times$) magnifications. 
This builds a visual search path across scales, which is then confirmed by senior pathologists.
Next, pathologists provide the rationale for each zoom-in decision based on the preceding field of view and provide detailed captions of pathological features for every ROI. Finally, we use the GPT-5.2 API to construct a cross-scale clinical synthesis from the zoom-in rationales and ROI captions. Pathologists then refine and confirm the fine-grained visual findings that explicitly link broad tissue architecture with localized cellular morphology.
These expert annotations serve as high-quality evidence anchors of the cross-scale search process, enabling clinically meaningful supervision for subsequent VQA curation and model training.

\subsection{Scoping Cross-scale Instruction with Leakage-Aware VQA}
Building on these annotations, we construct \dataset\ via a leakage-aware cross-scale curation pipeline that suppresses text-only shortcut solutions (Fig.~\ref{fig:main}a). The pipeline consists of three steps:

\noindent \textbf{Step 1: Scale-specific Feature Decomposition.} 
We decompose expert annotations into scale-specific evidence sets to reduce semantic leakage and linguistic shortcuts that can arise from full captions. To ensure clinical relevance, we define five cross-scale reasoning dimensions, namely \textit{Correspondence}, \textit{Confirmation}, \textit{Localization}, \textit{Explanation}, and \textit{Diagnosis}. These dimensions were designed and validated with pathologists to assess the multifaceted cross-scale knowledge used in practical clinical assessment. We also establish a set of initial constraints that guide the VQA generation process: (i) visual-grounding constraint that ties each question to the provided images, (ii) scale-dependency constraint requiring evidence from at least two magnification levels, and (iii) dimension-specific constraints to ensure diverse questions across reasoning dimensions for the same image set. 
These scale-specific feature sets and constraints are then provided as input to GPT-5.2 to generate the candidate MCQs.

\noindent \textbf{Step 2: Leakage-aware Screening via Text-Only Adversarial Inference.} 
To detect and eliminate text-only shortcuts, we introduce an iterative screening loop. For each candidate MCQ, we use Gemini 3 Pro and Qwen3-Max as text-only adversaries that receive only the question and options. If either model answers correctly, we analyze its rationale to diagnose leakage (e.g., linguistic cues, label priors, or unbalanced distractors) to update the constraints. We repeat this process until the adversaries can no longer reliably infer the correct answer without visual input.

\noindent \textbf{Step 3: Cross-scale MCQ Construction and Clinical Validation.} 
Using the refined constraints, we generate the final MCQs, each paired with multi-scale images (low/mid/high), a correct option, and a concise ground-truth rationale. We further post-process the dataset by merging questions across the five dimensions and balancing answer positions to reduce positional bias. Finally, pathologists validate the questions to ensure that correct answers are visually supported and distractors remain clinically plausible.

Together, this pipeline results in \dataset, a high-quality dataset that reliably measures cross-scale reasoning with 4,685 MCQs across 15 organs and 5 reasoning dimensions (Fig.~\ref{fig:main}b). By requiring coordinated multi-image, multi-magnification evidence integration, \dataset\ provides a principled foundation for training and evaluating pathology VLMs in a setting that better reflects real-world diagnostic workflows.

\subsection{Cross-scale Reasoning Optimization via RL}
We train \model\ on \dataset\ using Group Relative Policy Optimization (GRPO) ~\cite{deepseekr1} to improve answer selection for cross-scale pathology VQA. While standard supervised fine-tuning (SFT) can overfit to demonstration-style traces or surface correlations, RL provides outcome-driven feedback that encourages \model\ to learn decision rules that integrate evidence across magnifications, without requiring costly step-by-step human rationales. As illustrated in Fig.~\ref{fig:main}(c), given a multi-scale image set $I=\{I^{10\times}, I^{40\times}, I^{200\times}\}$, a question $q$, and options $\mathcal{O}$, the policy $\pi_\theta$ generates a structured response with reasoning (\texttt{<think>}) and a final answer (\texttt{<answer>}). We optimize $\pi_\theta$ with a sparse reward $R=\lambda R_{\text{acc}} + (1-\lambda)R_{\text{form}}$, where $R_{\text{acc}}=1$ if the selected option matches the ground truth (and $0$ otherwise), and $R_{\text{form}}$ enforces valid output formatting; we set $\lambda=0.8$. GRPO updates the policy by sampling multiple responses per question and computing group-relative advantages, eliminating the need for an explicit critic and encouraging responses that leverage cross-scale visual evidence to reach correct decisions.

\section{Experiments}
\subsection{Implementation Details}
\noindent \textbf{Training Setup.}  We patient-wise split \dataset\ into train/validation/test with 3{,}230/450/1,005 samples.  All multi-scale views and VQA samples derived from the same WSI are kept within the same split to avoid leakage. \model\ is initialized from Patho-R1-7B~\cite{pathor1}, a pathology VLM trained on large-scale pathology data but focused only on single-image analysis. This provides strong domain knowledge before adapting to our cross-scale setting. Based on UnPuzzle Pipeline~\cite{unpuzzle}, we train with GRPO for 500 steps on 8$\times$H100 GPUs.

\noindent \textbf{Benchmarks.}
We evaluate our model on two VQA settings: (1) \textit{Cross-scale multi-image:} \dataset-Test containing 1,005 VQA samples that require reasoning across magnifications. Performance is reported as accuracy across 5 clinically-aligned reasoning dimensions; (2) \textit{Single-image evaluation: } To assess model performance within the conventional single-image diagnostic paradigm, we utilize PathMMU~\cite{pathmmu}, an authoritative and large-scale public pathological VQA benchmark. We conduct a comprehensive evaluation across all PathMMU partitions (val/test-tiny/test), with a total of 10{,}387 VQA samples across 5 subsets, namely, Atlas, EduContent, PathCLS, PubMed, and SocialPath.

\noindent \textbf{Compared models.} We compare against representative models across domains: general VLMs (Qwen2.5-VL-7B~\cite{qwen25vl}, GPT-5.2~\cite{gpt}, Gemini 3 Flash~\cite{gemini}), medical VLMs (LLaVA-Med-7B~\cite{llavamed}, HuatuoGPT-7B~\cite{huatuogpt}, Lingshu-7B~\cite{lingshu}), and pathology VLMs (Quilt-LLaVA~\cite{quilt}, CLOVER~\cite{clover}, Patho-R1-7B~\cite{pathor1}).

\subsection{Comparison Results}

\begin{table}[t!]
  \caption{Accuracy on the cross-scale multi-image \dataset-Test benchmark. Results are reported across five cross-scale reasoning dimensions (Correspondence, Confirmation, Localization, Explanation, Diagnosis) and averaged (AVG). The best results are highlighted in \textbf{bold}.}
  \label{tab:tripletvqa}
  \centering

  {\fontsize{7}{9.5}\selectfont

  \begin{tabular}{lcccccc}
    \toprule
    & \textit{\textbf{Corresp.}} & \textit{\textbf{Confirm.}} & \textit{\textbf{Localiz.}} &
      \textit{\textbf{Explan.}}  & \textit{\textbf{Diagno.}}   & \textbf{AVG} \\
    \midrule

    \multicolumn{7}{l}{\textbf{General VLM}} \\
    \hdashline\addlinespace
    Qwen2.5VL-7B   & 41.79 & 47.26 & 71.14 & 44.28 & 63.68 & 53.63 \\
    Gemini 3 Flash & 48.26 & 58.71 & 71.64 & 53.73 & 72.14 & 60.90 \\
    GPT-5.2        & 47.76 & 59.70 & 74.13 & 45.77 & 65.17 & 58.51 \\
    \midrule

    \multicolumn{7}{l}{\textbf{Medical VLM}} \\
    \hdashline\addlinespace
    LLaVA-Med-7B   & 25.87 & 16.92 & 28.36 & 20.90 & 24.88 & 23.38 \\
    HuatuoGPT-7B   & 37.81 & 45.77 & 65.67 & 46.77 & 55.72 & 50.35 \\
    Lingshu-7B     & 47.26 & 63.68 & 62.69 & 48.26 & 61.19 & 56.62 \\
    \midrule

    \multicolumn{7}{l}{\textbf{Pathological VLM}} \\
    \hdashline\addlinespace
    Quilt-LLaVA    & 32.34 & 14.43 & 45.27 & 30.35 & 29.35 & 30.35 \\
    CLOVER         & 37.31 & 61.69 & 73.13 & 46.27 & 65.77 & 56.82 \\
    Patho-R1       & 31.84 & 40.30 & 59.70 & 56.22 & 68.66 & 51.34 \\

    \rowcolor{pinky}
    \textbf{\model} & \textbf{80.60} & \textbf{89.05} & \textbf{84.58} & \textbf{76.12} & \textbf{84.08} & \textbf{82.89} \\
    \addlinespace[2pt]\hline\hline\addlinespace[2pt]

    \multicolumn{7}{l}{\textbf{Ablation Study}} \\
    \hdashline\addlinespace
    \rowcolor{pinky!30}
    SFT            & 69.65 & 83.58 & 78.11 & 55.22 & 64.68 & 70.25 \\
    \rowcolor{pinky!30}
    SFT + RL       & 72.64 & 86.57 & 79.10 & 57.71 & 68.16 & 72.84 \\
    \bottomrule
  \end{tabular}
  }
\end{table}

As shown in Table~\ref{tab:tripletvqa} and Fig.~\ref{fig:main}(d), \model\ establishes a new state-of-the-art for cross-scale reasoning, achieving an average accuracy of 82.89\% on \dataset-Test. 
Compared to its base model, our approach yields a remarkable improvement across all five reasoning dimensions despite the relatively small-scale training data. 
Notably, although trained only on cross-scale multi-image questions, \model\ also improves performance on the single-image PathMMU benchmark
(Table~\ref{tab:pathmmu}). 
Our model's performance even surpasses or approaches the expert performance reported in the original PathMMU paper for most of the subsets, such as in EduContent and Atlas. We also observe a modest drop on PathCLS, which is closer to label-centric single-image classification than morphology-grounded reasoning, suggesting a possible trade-off between cross-scale evidence integration and category-recognition priors. These results demonstrate that cross-scale reasoning supervision strengthens the model’s pathological understanding and transfers beyond the training setting to improve on conventional single-image pathology VQA. 

\begin{table*}[!t]
	\caption{Overall results of models on the PathMMU validation and test set. The best model performance in each column is highlighted in \textbf{bold}. Results marked with $\dagger$ are taken directly from the original PathMMU paper.}
    \label{tab:pathmmu}
	\centering
    \resizebox{\textwidth}{!}{%
		\begin{tabular}{@{}lccccccccccccc@{}}
			\toprule
			\textbf{} & \multirow{2}{*}{\textbf{\begin{tabular}[c]{@{}c@{}}Val \\ overall\end{tabular}}} & \multicolumn{2}{c}{\textbf{Test overall}} & \multicolumn{2}{c}{\textbf{\textit{PubMed}}} & \multicolumn{2}{c}{\textbf{\textit{SocialPath}}} & \multicolumn{2}{c}{\textbf{\textit{EduContent}}} & \multicolumn{2}{c}{\textbf{\textit{Atlas}}} &\multicolumn{2}{c}{\textbf{\textit{PathCLS}}} \\ 
			&- & Tiny  & All  & Tiny  & ALL & Tiny  & All & Tiny  & All  & Tiny  & All  & Tiny  & All\\\midrule
			
			\color{gray} Random Choice$^\dagger$ & \color{gray} 24.6 & \color{gray} 22.1 & \color{gray} 23.7 & \color{gray} 22.1 & \color{gray} 25.1 & \color{gray} 25.5 & \color{gray} 26.5 & \color{gray} 25.5 & \color{gray} 26.0 & \color{gray} 19.7 & \color{gray} 23.0  & \color{gray} 15.3 & \color{gray} 16.3\\ 
			\color{gray} Frequent Choice$^\dagger$ & \color{gray} 27.5 & \color{gray} 27.7 & \color{gray} 25.5 & \color{gray} 28.8 & \color{gray} 26.1 & \color{gray} 27.7 & \color{gray} 26.7 & \color{gray} 29.8 & \color{gray} 26.5 & \color{gray} 28.4 & \color{gray} 27.5 & \color{gray} 22.0 & \color{gray} 21.0  \\
			\rowcolor{aliceblue!60}  Expert performance$^\dagger$ & - &  71.8 &  - &  72.9 & -& 71.5 &  - &  69.0 &  - & 68.3 &  - & 78.9 &  - \\
			\midrule
			\multicolumn{12}{l}{\textbf{General domain VLM}} \\ 
			\hdashline\addlinespace
			Qwen2.5-VL-7B      & 44.3 & 48.2 & 44.4 & 53.4 & 50.5 & 50.0 & 47.7 & 59.6 & 48.9 & 48.6 & 46.8 & 29.4 & 28.0 \\
			Gemini Pro Vision$^\dagger$  & 41.9 & 42.8 & 42.7 & 43.8 & 44.9 & 42.4 & 42.0 & 43.5 & 43.7 & 49.5 & 49.4 & 32.8 & 34.7   \\
			GPT-4V$^\dagger$          & 49.3 & 53.9 & 49.8 & 59.4 & 53.5 & 58.7 & 53.9 & 60.4 & 53.6 & 48.1 & 52.8  & 36.2 & 33.8\\	
            \midrule
			\multicolumn{12}{l}{\textbf{Medical VLM}} \\ 
            \hdashline\addlinespace
            LLaVA-Med-7B & 17.5 & 22.2 & 22.7 & 24.6 & 25.1 & 23.4 & 21.1 & 25.1 & 23.6 & 17.8 & 24.5 & 20.3 & 19.2 \\
            HuatuoGPT-V-7B & 43.2 & 43.8 & 41.2 & 49.1 & 47.7 & 52.3 & 46.0 & 52.9 & 46.1 & 44.7 & 46.9 & 19.8 & 19.2 \\
            Lingshu-7B & 51.9 & 56.0 & 53.7 & 60.5 & 57.6 & 61.6 & 58.1 & 69.4 & 58.7 & 58.1 & 62.5 & 30.4 & 31.6 \\	 \midrule
			\multicolumn{12}{l}{\textbf{Pathological VLM}} \\ 
            \hdashline\addlinespace
             Quilt-LLaVA & 33.8 & 32.3 & 31.9 & 30.3 & 34.2 & 26.8 & 34.6 & 35.7 & 33.3 & 41.8 & 33.3 & 27.1 & 24.0 \\
            CLOVER & 53.1 & 60.6 & 56.6 & 68.0 & 59.8 & 67.6 & 60.5 & 68.6 & 62.2 & 62.0 & 64.1 & 36.7 & 36.6 \\
            Patho-R1 & 62.3 & 64.8 & 62.6 & 68.7 & 64.9 & 63.9 & 65.7 & 71.0 & 66.1 & 78.4 & 73.5 & \textbf{41.8} & \textbf{42.7} \\
            \rowcolor{pinky} \textbf{\model}& \textbf{62.4} & \textbf{66.2} & \textbf{63.8} & \textbf{71.2} & \textbf{67.3} & \textbf{67.6} & \textbf{67.6} & \textbf{74.9} & \textbf{69.2} & \textbf{79.3} & \textbf{76.2} & 37.9 & 38.5 \\
            \addlinespace[2pt]\hline\hline \addlinespace[2pt]
            \multicolumn{7}{l}{\textbf{Ablation Study}} \\
            \hdashline\addlinespace
            \rowcolor{pinky!30} SFT & 47.8 & 44.0 & 43.3 & 59.1 & 52.1 & 57.9	 & 53.3 & 56.9 & 51.6 & 43.3 & 47.7 & 22.0 & 19.6 \\
            \rowcolor{pinky!30} SFT + RL & 48.3 & 45.2 & 44.1 & 57.7 & 52.7 & 56.0 & 52.8 & 57.6 & 52.9 & 46.6 & 48.1 & 23.7 & 19.7 \\
			\bottomrule
		\end{tabular}
	}
\end{table*}

\subsection{Ablation Study: Activating Generalized Pathological Reasoning via RL}
To gain deeper insights into the training dynamics and the necessity of our outcome-driven optimization, we compare our GRPO-based RL against SFT-only and SFT+RL paradigms. To enable SFT, we constructed an instruction-tuning set comprising 4,475 cross-scale VQA samples derived from 93 WSIs, pairing each question’s rationale with its cross-scale caption. All variants share the Patho-R1-7B backbone and are evaluated uniformly. Results on cross-scale (Table~\ref{tab:tripletvqa}) and single-scale benchmarks (Table~\ref{tab:pathmmu}) reveal a clear dynamic: RL-only consistently outperforms both SFT-only and SFT+RL. We observe that SFT, while improving cross-scale accuracy compared to baseline, suffers a noticeable drop on single-scale benchmarks. This indicates a strict trade-off where forcing the model to mimic instruction-style rationales leads to catastrophic forgetting of its pre-trained single-scale representations. SFT+RL mitigates this degradation but still falls short of pure RL. This demonstrates that pure RL avoids the bottleneck of linguistic imitation: rather than training the model to reproduce human-written rationales, RL directly optimizes answer selection from cross-scale visual evidence via outcome rewards. Our observation aligns with \textit{SFT Memorizes, RL Generalizes} ~\cite{sftrl}, which similarly reports that SFT can overfit to demonstration traces while outcome-based RL yields stronger generalization. Together, these results demonstrate the strength of our RL-centric approach as a more effective and annotation-efficient strategy for cross-scale pathology VQA in the low-data setting. 
\section{Conclusion}

In conclusion, we present the first cross-scale training and evaluation paradigm for pathological VLMs, framing diagnosis as coordinated reasoning across magnifications.
We release a high-quality dataset \dataset, constructed with cross-scale captions and a leakage-aware VQA curation pipeline that suppresses text-only shortcuts and enforces multi-view evidence integration.
We further introduce \model, a data-efficient RL-trained model for cross-scale reasoning, achieving state-of-the-art performance on both multi-magnification and standard single-image pathology VQA benchmarks.
These gains improve diagnostic robustness at both single and multiple scales, underscoring the practical value of cross-scale supervision.

\begin{credits}
\subsubsection{\ackname} This work was supported by the Ministry of Education Tier 1 grant, Singapore (24-1250-P0001), and the Ministry of Education Tier 2 grant, Singapore (T2EP20224-0028). This work was powered by the UnPuzzle \& PuzzleCloud Platform (https://puzzlelogic.com/unpuzzle) and supported by PuzzleLogic Pte Ltd, Singapore.

\subsubsection{\discintname}
The authors have no competing interests to declare that are relevant to the content of this article.
\end{credits}

\bibliographystyle{splncs04}
\bibliography{Sections/Reference}

\end{document}